# Extensible Grounding of Speech for Robot Instruction

**Jonathan H. Connell**
IBM T.J. Watson Research Center
Yorktown Heights, New York USA
*jconnell@us.ibm.com*

*Abstract*

*Spoken language is a convenient interface for commanding a mobile robot. Yet for this to work a number of base terms must be grounded in perceptual and motor skills. We detail the language processing used on our robot ELI and explain how this grounding is performed, how it interacts with user gestures, and how it handles phenomena such as anaphora. More importantly, however, there are certain concepts which the robot cannot be preprogrammed with, such as the names of various objects in a household or the nature of specific tasks it may be requested to perform. In these cases it is vital that there exist a method for extending the grounding, essentially "learning by being told". We describe how this was successfully implemented for learning new nouns and verbs in a tabletop setting. Creating this language learning kernel may be the last explicit programming the robot ever needs – the core mechanism could eventually be used for imparting a vast amount of knowledge, much as a child learns from its parents and teachers.*

## 1. Introduction

Suppose you buy a general fetch-and-carry robot from Sears and take it home. It will come with all sorts of extraordinary innate abilities like collision avoidance, optimal navigation, object recognition, grasp planning, and visual search routines. However, to customize it to do such valuable tasks as serving beer, it needs to be told something about your environment. You might take it to the kitchen, show it the fridge, and point out where you keep the beer. You might then tell it what was expected of it in response to commands such as "Get me a Dos Equis".

Note that here the user is not doing heavy-duty programming at the level of setting joint torques or specifying convolution masks. In many ways this training is more like working with a simple macro scripting language – you define a few new items in terms of the system's base types, and provide rewrite rules for translating commands into a parameterized sequence of built-in actions. While this is not the only way to customize a robot, nor the only use for natural language interaction, the two seem to be a good match.

### 1.1 Eldercare as a Domain

A robot that could perform simple fetch-and-carry tasks would have many potential applications (besides serving beer). One of the domains in which it might be most valuable is eldercare, and this is also where a natural, customizable interface is most desirable. What we would like is essentially a service dog with more language and less slobber. Consideration of this domain has influenced much of the design and functionality of our robot.

In an eldercare scenario a human aide would still likely be required for activities such as dressing, bathing, and food preparation, yet there are a number of simpler activities that could be economically performed by a low cost robot. These include things like picking up a dropped phone, since many elders have difficulty getting out of chairs to bend over. Another example would be to bring a blanket from a different room. Carrying a bulky object potentially leads to balance problems and tripping hazards. Still another example would be to retrieve the book that was being read yesterday. The senior may have forgotten exactly where it was left, and clumping around the house with a walker takes a surprising amount of time. A robot that could help with these menial chores would boost the senior's feeling of control and allow them to stay in their homes longer (likely at lower cost than at an assisted living facility).

A primary technical issue that must be addressed with eldercare assistance is how to tell the robot what to do. The current generation of senior citizens is not comfortable with tablets, keyboards, styli, PDAs, or Bluetooth headsets. Moreover, all of these accessories are just one more thing to drop or misplace. The most human-friendly interface is direct speech using a audio pickup on the robot itself. The trick then is interpreting the spoken commands robustly. In addition, a particular home may have locations, like the "solarium", or objects, like "my favorite cup", which cannot be known *a priori* and hence cannot be preprogrammed into the robot. Thus it would be convenient if the robot could just be shown such places and objects and learn whatever models it needed automatically. In addition there may be activities such as "tidy up the nightstand" that are specific to an individual. Again, being able to learn these things on the fly given verbal (and perhaps gestural) guidance would be a benefit.

Of course ours is not the first attempt at a home robot or the first mobile manipulator. There is the impressive PR2 from Willow Garage (Chitta et al. 2012) which can do things like fold towels (but slowly, and for $400K). HERB, developed at CMU (Srinivasa et al. 2010) is also intended to perform household tasks, but currently requires environmental modifications for its vision system. Then there is El-E from Georgia Tech (Choi et al. 2009) that was specifically created to retrieve objects for disabled persons. However, none of these robots are designed around a speech interface – to change their actions you either completely change their programs or you configure options in a GUI. Other robot such as Carl (Lopes & Teixeira 2000) and Cosero (Stuckler et al. 2012) can take speech input, but require a handheld or headset mike. Furthermore, in general these robots are not intended to learn in the field from user interaction. Instead they have various preprogrammed competencies, object models, and environmental maps which are developed offline.

Other work has addressed language-based learning. Much of this, however, has started at a very low level. Steels (1998) looks at the emergence of a private language between cooperating agents while Roy (2003) attempts to directly associate acoustic fragments with visual fragments. What we believe is more useful is to stick with a human language and just attempt to find suitable bindings for a few unknown words. This is akin to the approach taken in HAM (Peltason et al. 2009) for learning place names. Similarly, procedure learning is often attempted through trial-and-error experimentation (Siskind 2001) or using the impoverished feedback of reinforcement learning (Breazeal et al. 2006). Yet explicit macro definitions or verbal scripting is often faster and more effective in practice.

## 1.2 Language and Learning

It is important to recognize that language is not just a channel for imparting commands, it is also an invaluable tool for teaching. For instance, once the robot is up and running one can not just say "Bad dog!" and expect it to figure out what it did wrong. For one thing, the search space of alternative actions is apt to be huge, leading to a lot of trial and error before the robot gets it right. This will yield unsatisfactory performance in the meantime and an impression of an excessively obtuse servant. More importantly, however, the bad-dog response is ridiculously uninformative since the user usually knows exactly what the robot did wrong, or can quickly determine where the difficulty lies.

Generally, language is useful during learning because it allows one to direct the robot's attention precisely and efficiently. Suppose one just stopped the robot and said "supporting shelf". The robot now has to guess, first of all, whether some sort of command is being issued or a new item is being taught. It then has to figure out if "supporting shelf" is some sort of an action it supposed to learn, or if it is a spatial location, or even some kind of object. While pure induction might be able to sort out all these ambiguities given enough examples, the user is likely to get frustrated long before this occurs.

Suppose instead the user says "This LOCATION is a supporting shelf" thereby simultaneously denoting a spatial region and ascribing to it a previously known semantic type. From the type the robot can now infer that certain of the item's properties are important, such as its position relative to some reference, whereas as other properties, such as its color, are likely to be irrelevant. In concert with the spatial bounds, this can greatly reduce the dimensionality of the description space and make induction far more tractable. Furthermore, critical features such as "It must be level horizontally" can be communicated directly, something that might be difficult (or at least tedious) to infer from examples alone.

Even richer feedback can be provided through a debugging tool such as a stack prober. To implement such a facility the robot needs to be able to answer "why" it did a particular action. As in expert systems, this can be done by plugging the information from various call levels into standard pseudo-English response templates. In addition to straight dependency-directed reporting, it should also be possible to access the "comments" included during programming as a justification for the step from a higher authority, e.g. R: "What kind of beer do you want?", U: "I don't know, surprise me ... Aaack, why did you dump beer in my lap?", R: "To surprise you.", U: "Yes, but WHY?", R: "Because Joe told me this would be surprising."

Once the source of a difficulty has been pinpointed there are several things that might be done to correct it. Again, a verbal interface to at least the simplest level of an editing facility would be desirable. Often the confusion will involve the mis-identification of an object. This can typically be cured by adding or deleting conditions from a rule: "Show me a beer bottle ... No, a beer bottle has to be brown." or, conversely, U: "Why isn't this a beer bottle?" R: "It is not upright.", U: "Oh, that's not important."

Other times, extra meta-control information may need to be supplied (e.g. to resolve rule conflicts in a production system such as ACT* (Anderson 1983)). Suppose the robot was asked to bring a beer from the kitchen to the living room. If it knew the spatial layout of the house, its path planner might decide that proceeding directly through the door connecting the two rooms was the most efficient route. Yet the user might object to this and attempt to reprogram the robot through a dialog such as: U: "How ELSE could you get to the living room?", R: "Through the hallway.", U: "That is better because you won't block the TV." Here a preference is set up for selecting a route, along with a comment-like justification (which the robot merely needs to parrot later).

## 1.3 Cultural Bootstrapping

In the larger Artificial Intelligence context, language and verbal learning are of paramount importance. Part of the feeling of aliveness comes from the responsiveness of a creature with a reasonably deep perception of its environment. However, even humans from a different society can be successfully demonized as "sub-human" if you cannot understand what they say. Moreover, if a person suffers amnesia and forgets his family and culture, in many ways he has been erased as the person he previously was. Thus it seems essential for a true AI to have the ability to learn and remember patterns of cultural interaction.

More pragmatically, the bulk of human-level intelligence is due to acquired cultural knowledge, at least observationally. Language is both the key for learning this and the medium by which proficiency in culture is evidenced (Vygotsky 1962). It is a sort of mind-sharing "telepathy" that makes an agent part of a cultural super-organism and allows its peers to "program" it to act in certain ways in specific situations (Skinner 1957). Tellingly, much of what we consider knowledge or competence is simply transmitted culturally. No one figures out how to cook macaroni and cheese by experimentation – some other person tells you how to do it. If robots are ever to be perceived as sentient it seems crucial that they also be able to learn in this manner and thus partake of the rich prevailing culture which underpins much of "human-ness".

If language is so important, how can a robot (or human) acquire it? At the bottom, language can be viewed as a general-purpose scripting system that invokes sensory/motor specific subroutines as needed. One way to become proficient in a language is to build an interpreter that allows the agent to receive descriptions and instructions via language then evaluate these structures to operationalize the knowledge. The starting point for this is to ground the fragmentary acoustic utterances in sensory and actuator experiences (Roy 2003, Peltason et al. 2009). As a prerequisite the system must have some innate mechanisms of parsing the world into physical objects and temporal events. It also needs some reflexes which will exercise all the actuator options, and an initial good/bad reinforcement signal as feedback. An intrinsic means to guide attention is also very useful. All these abilities can be bootstrapped to higher levels of sophistication with a method like clicker training (i.e. operant conditioning with staging and shaping (Pryor 1984)).

Once a basic level of linguistic grounding has been achieved, learning new concepts and tasks becomes much easier. For instance, one can say "No, this is a moth not a butterfly. Look at its fuzzy antennae." This explanation is much simpler and faster than showing the agent tens of contrasting examples and having it guess which features are relevant to the classification.

Similarly, language can be used to directly impart routines for accomplishing sequential tasks, such as opening a jar of pickles. Instead of letting the agent fiddle with it for hours and shouting "Hurray!" when the top finally pops off, language allows the teacher to say "Hold the jar in your left hand, grasp the top with your right hand, and twist hard". If this sequence of verbal directions is memorized and played back through the interpreter, the agent has essentially learned a plan for how to do the task in one shot (perhaps without ever actually performing it!).

Although our language interpreter is built with conventional technologies, consider a Turing machine by analogy. At its heart there is an FSM which, in itself, is not so interesting. Yet having something like this allows the creature to manipulate the "tape" of culture and thus greatly expand its capabilities.

## 2. Grounding Substrate

The driving force for the robot system detailed in the following sections is multimodal instructional dialog. We want the user to be able to describe a task, through a combination of speech and gesture (multi-modal), and then have the system successfully accomplish the task. If the robot is unsure about some aspect of the task, it should ask clarifying questions (dialog). In addition, we want the system to be able to learn about new objects and new procedures to enable a "verbal programming" facility (instruction).

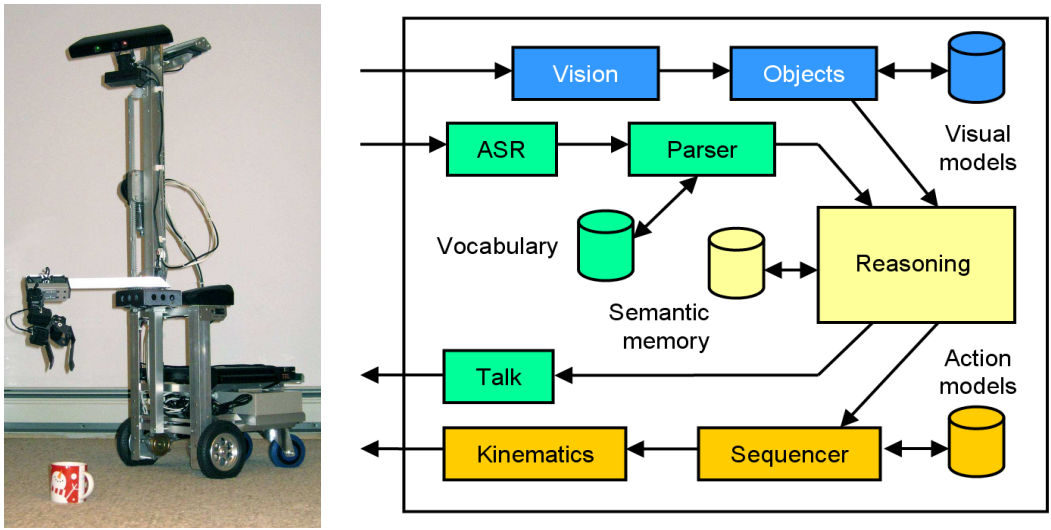

**Fig. 1.** The robot was designed for eldercare applications and can interpret spoken commands. It also has the ability to learn new nouns and verbs from interaction with its user. There are several stages to the interpretation and learning process, and separate updatable databases associated with each.

Fig. 1 shows a block diagram for our system and the physical robot it controls. We call the system the Extensible Language Interface (ELI): a speech guided mobile robot that can learn new nouns and new verbs based on user instruction. The structure of the robot itself follows from the eldercare application: far-field speech recognition, simple manipulation, ability to reach both the floor and countertops, and a small footprint for navigating within an apartment. However, to reduce the degrees of freedom to be controlled, the actual experiments here were performed by removing the arm and camera from the robot and mounting them on a tabletop.

## 2.1 Object Finding

Since our example tasks all concern tabletop manipulation, it is important for the robot to detect objects. To do this it looks for "obvious" objects: things that are clearly separated from each other on some working surface. For simple 2D color images we use a heuristic approach and look for things that are a different color than the table. To do this we have to first determine what the color of the table is. Fig. 2 shows some of the steps in the processing. Given an input image (top left) we first enhance its overall colorfulness (top right) by multiplying the RGB values at each pixel by a single factor ($< 5$) such that at least one of the color channels (R, G, B) achieves the maximum value of 255. This boosts the saturation and intensity of the image, but leaves the hue values unchanged. We then compute color difference images, which are shown in pseudo-color below. There is R-G, the red channel minus the blue channel (lower left), and Y-B, the sum of the red and green channels minus twice the blue channel (lower right).

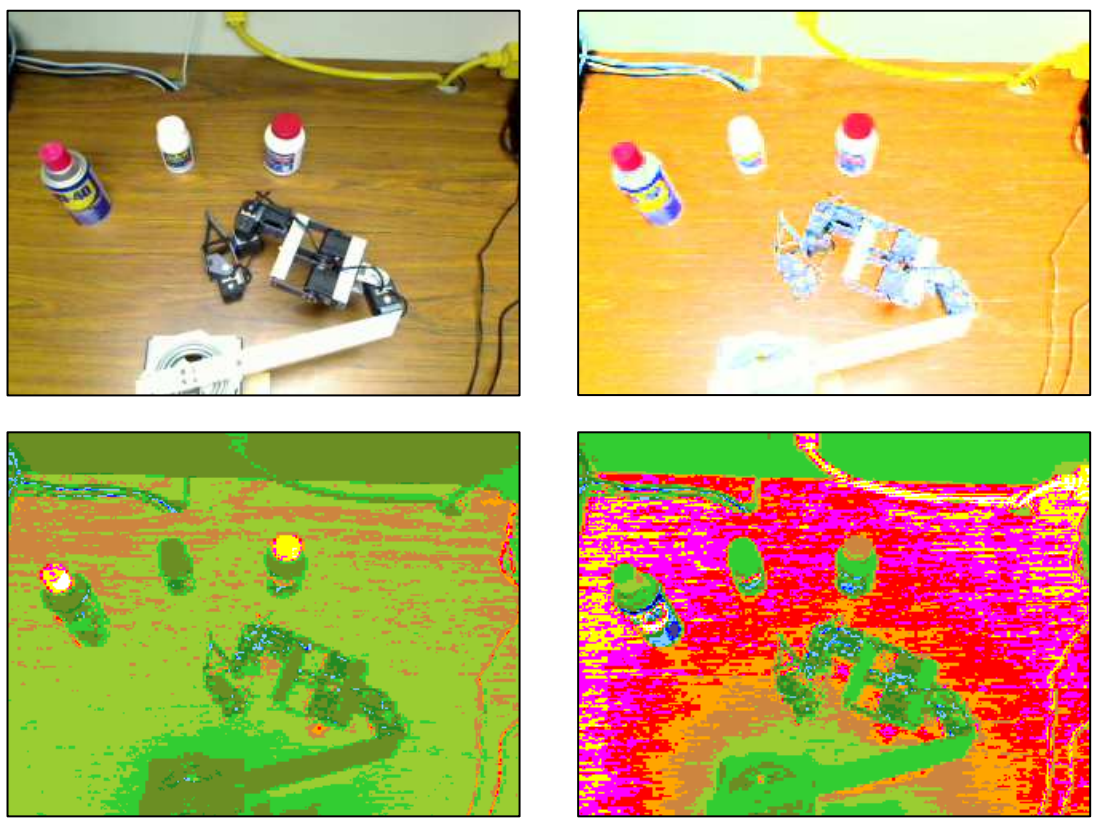

**Fig. 2.** Finding objects on a table requires determining the color of the table. We first boost the colorfulness of the scene (top) then analyze it in terms of RGB channel differences at each pixel (bottom).

The robot then histograms the opponent color values in a region which is essentially a horizontal band covering the center third of the image. This is the likely location for the bulk of the table, ignoring any floor which might be visible below and any wall that might be visible above. Fig. 3 shows the RG and YB histograms (left). From these the robot finds the peak value and sets bounding thresholds based on where the peak has dropped to a given fraction of its maximal value. These thresholds are then used to create a binary version of the input image where table pixels are white and non-table pixels are black (right). The black areas are where the pixel color is either outside the red-green limits or outside the yellow-blue limits for the inferred table color.

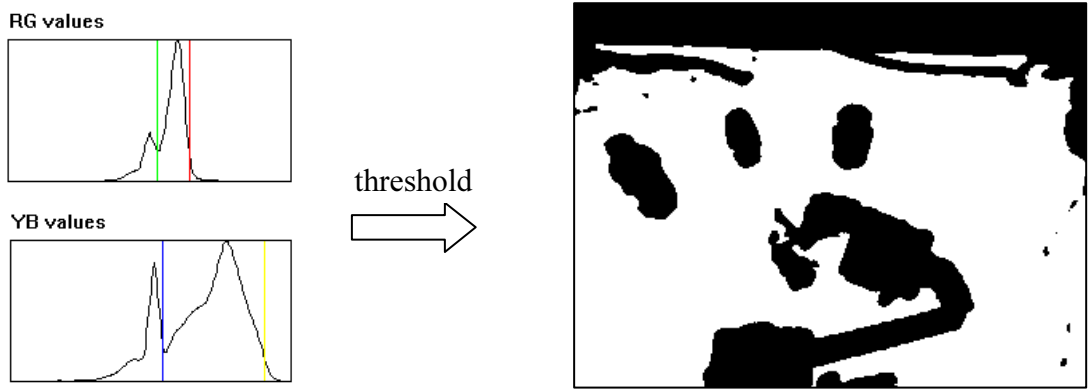

**Fig. 3.** Potential objects are regions with a color different than the table. The thresholds are determined by finding peaks in the RG and YB histograms, which have been formed using pixels likely to have come from the table.

A similar method can be used with the depth camera on the large robot, as shown in Fig. 4. However instead of modeling the table in terms of a dominant color, it is modeled as a 3D plane. The left side of this figure shows intensity-coded depth – the farther away objects are, the darker they are (full black marks regions where the depth could not be determined). As can be seen, there is a gradual shading from light at the bottom of the image to darker at the top. We split the image into a series of vertical strips, fit a 2D line to the gray values within each strip, then combine compatible line to yield an overall 3D plane estimate. The right side of the figure shows the pixels compatible with this flat surface (green). Again, deviations from this model are considered potential objects after suitable noise reduction and morphology operators are applied.

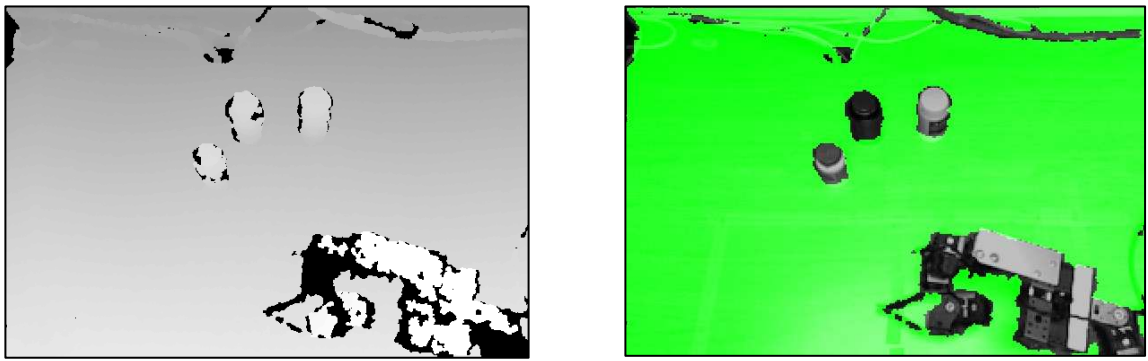

**Fig. 4.** Objects can also be found by looking for geometric deviations from a supporting surface. The left image shows a 3D range map from the onboard Kinect sensor. On the right we fit a plane to the depth points then isolate regions which protrude.

## 2.2 Object Properties

Using the rough binary map computed from either color or depth, a further series of operations is needed to cleanly define objects. First, the jagged edges of the mask are smoothed then a connected component analysis is performed to find the largest contiguous table area. A second connected component analysis is then performed on the "holes" in this area to yield potential objects. The result is shown on the right side of Fig. 5.

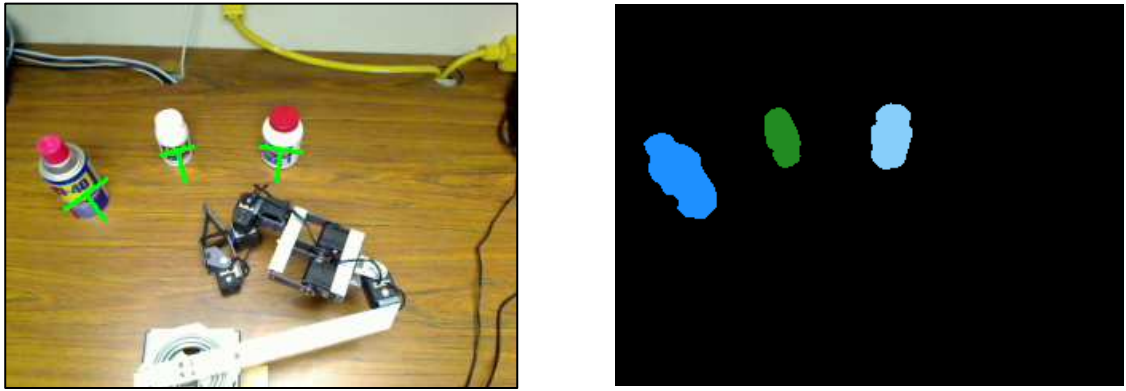

**Fig. 5.** A variety of morphology and connectivity operations are performed to find the basic objects (right). Afterward a surface position and grasp approach direction are computed for each item (left).

A secondary result of the connected component analysis is the computation of an axis of minimal inertia. This generally follows the axis of elongation of the binary blob, and is indicated by the direction of the base of the green "T" shown for each object on the left of Fig. 5. This gives a hint of a good approach direction for grasping the object with the robot arm. This axis is also used to infer the point on the table surface associated with the center of the base of the object, which is needed to guide the arm to it. The base point is found by first computing a width projection along the axis for each object, then figuring out the characteristic width near the lower portion of the blob. The table point is then half this width up from the lowest axial pixel in the blob, as indicated by where the base meets the top of the green "T".

We can now ground spatial and size parameters of each object. Having reduced them to a point, it is now easy to determine the "leftmost" and "rightmost" objects. Note that this can also be done relative to some other selection criteria such as "only red things", which serves to reduce the number of objects under consideration. Determining the one "in the middle" requires finding the leftmost and rightmost, then picking the object closest to the average of these two image positions. The concepts of "biggest" and "smallest" are based on the pixel count of each blob. This is directly related to their true size since the camera is looking nearly straight down and is far enough away for a largely orthographic projection. Note that we have found size to be better as a ranking criterion (i.e. comparative degree) rather than a value to threshold on. Thus "the big white object" is the biggest object which is white although, in actuality, the object itself may be fairly small in an absolute sense.

The binary mask for each object is then used to compute a color representation for it. As shown in Fig. 6 we take the pixels for an object (not bright blue) and shrink the mask a little since the borders are often corrupted by surrounding colors. Then, for each pixel, we classify it into one of 9 semantic color classes: red, orange, yellow, green, blue, violet, black, gray, or white. These bins are based on the HSI transform of the original RGB image. Pixels that are very bright or very dark in intensity are mapped to white (W) and black (K), respectively. Pixels in the middle range with low saturation are deemed gray (N). The remaining pixels with moderate intensity and high enough saturation are broken up between the true color categories (ROYGBV) based on hue thresholds. The second image in Fig. 6 shows the object recolored with these canonical values.

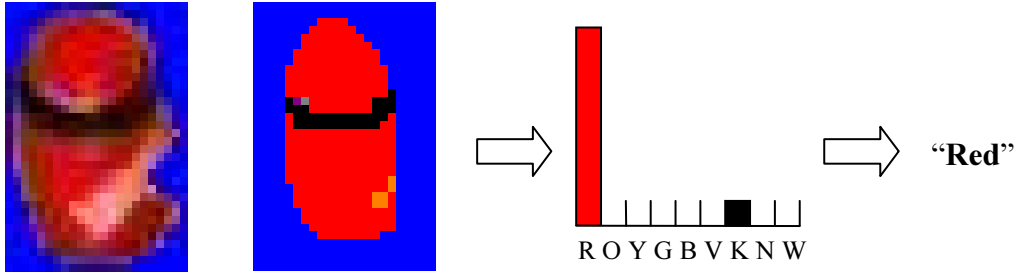

**Fig. 6.** Color is interpreted by defining 9 discrete categories in HSI space and histogramming the object pixels.

This image with the actual pixel classes is not saved. Rather, we histogram the object region to find out how many pixels there are of each type. For instance, as shown to the right in Fig. 6, roughly 80% of the pixels are "red" and about 20% are "black". This histogram is used for matching to stored object models and also further reduced to yield a dominant color, such as "red". This is useful when the robot is asked "what color is that object?". Sometimes, however, objects have a more varied mix of colors, so the answer might be "white with a little red" for a Tylenol bottle. A single color is chosen if its bin is twice as large as any other bin. If not, other colors that have bins that are more than 30% of the dominant color bin are added to the textual description. These expanded color descriptions can also be used to aid speech interpretation. Suppose that on the table there are a bunch of mostly white bottles with variously colored tops. Asking for the "red one" can then select the item with at least some red in it, rather than coming up empty handed.

### 2.3 Gesture Recognition

Another important basic capability is to understand human gesture. To do this we assume the user's hand is moving and use background subtraction to find it. At opportune points during the conversation when there is likely no motion present, the robot snaps a picture of the scene. Then it continually compares this "background" to the current image to find regions that have changed, presumably due to some sort of motion. For the scene on the right of Fig. 7, the difference regions are shown in white on the left side. Once again, this same algorithm for hand detection can be applied even more easily and robustly to depth data.

Still, the binary mask is not directly usable in its graphical form. We need something more like a mouse pointer position and a mouse click in order to indicate objects. To convert the bitmask to a position we find the bounding box of the motion region (red lines on left) and monitor how far the corner of this region (green cross) gets from the lower left corner of the image (the human's hand usually enters from this direction). When the tracked cross position stops changing, or shrinks back toward the image corner, we declare a mouse click. This can be then be used to find the nearest object based on its axis parallel bounding box (purple rectangle on right) and thus bind the referent for "this object".

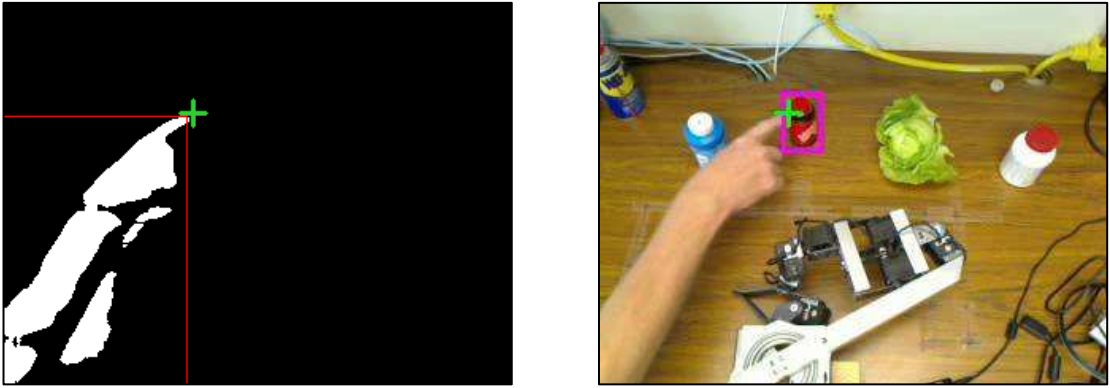

**Fig. 7.** Gesture recognition is implemented by using background subtraction to track the user's hand. The most extremal portion of this mask (left) selects one of the objects previously identified (right).

Similar processing is used to sequence hand-off operations. The left side of Fig. 8 shows the same background subtraction and point tracking as used before. However in this case we look for a "click" in a specially designated transfer zone (purple square in right panel) where the robot brings objects to be given to the user. When this happens the robot releases its object. It then waits for a second click in the transfer zone, presumably signaling that the user has returned the object in a suitable grasping position. At this point it closes it hand on the object and returns it to some previously determined home location.

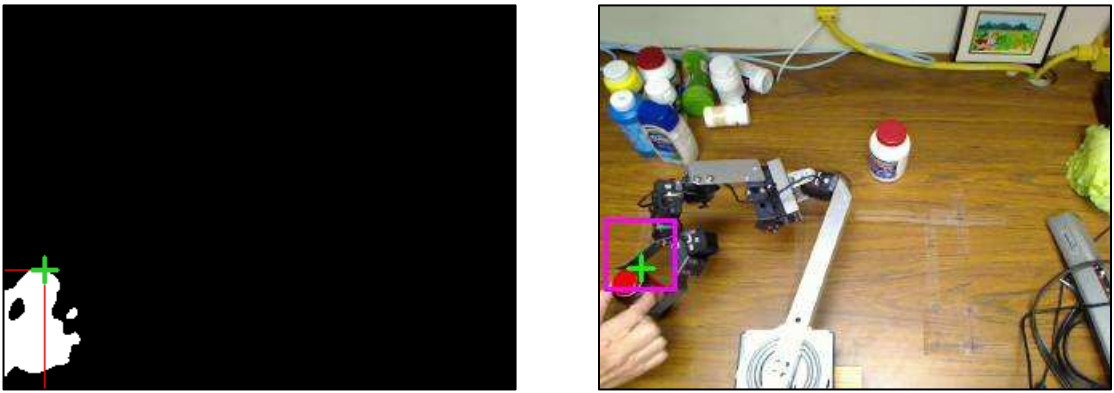

**Fig. 8.** User motion detection for object handoff works in a similar manner to pointing detection.

## 2.4 Speech Interpretation

For speech recognition we use an Acoustic Magic VT2 far-field array microphone. This input device not only performs beam forming using 6 microphone capsules, but also does automatic beam steering to pickup the dominant sound source. Speech interpretation is performed with the Microsoft ASR Engine in Windows XP (or Windows 7), although we have also successfully used the IBM Attila engine (Soltau et al. 2010). Far field speech is notoriously hard for machines to understand and prone to false detections. To improve recognition performance we use a speaker-dependent voice model, constrain the utterances with a command-and-control grammar, and only respond to requests prefixed (or suffixed) by the robot's name.

```
=[top_level]                          =[hand_grab]              =[COLOR]
<attn> (<intro>) <request> *          grab                      <red>
* <request> (<intro>) <attn>          grasp                     <orange>
<desc> <FACT> <pnoun> <attn>          lift                      <yellow>
                                      touch                     <green>
=[attn]                               pick                      <blue>
Eli                                   pick up                   <purple>
robot                                 select                    <black>
                                                                <gray>
=[intro]                              =[desc]                   <white>
please                                <np> (<pp>)
first                                                           =[blue]
next                                  =[np]                      blue
                                      <PRON>                    dark blue
=[request]                            <POINT> <obj>             light blue
<QUERY> <desc>                        (<det>) (<SIZE>) (<COLOR>) <obj>
<CMD> <desc>                          (<det>) (<POSITION>) (<COLOR>)   =[obj]
<ACT-0>                               <obj>                     (<measure>) <NAME>
<ACT-1> <desc>                                                  <REF>
<learn>                               =[det]                    object
                                      the                       objects
=[CMD]                                a                         thing
<hand_indicate>                       an                        things
<hand_select>                                                   bottle
<hand_grab>                           =[POINT]                  bottles
<hand_give>                           this
                                      that                      =[DICT]
=<FACT>                                                         +
is (<det>)                            =[pnoun]
is called                             <NAME>                    =[NAME]
is named                              <DICT>                    Tylenol
```

**Fig. 9.** This is part of the grammar used for speech interpretation and learning.

An example semantic grammar is shown in Fig. 9. Here there are a number of rules prefixed by "=" that offer several valid expansions for each non-terminal. Elements in parenthesis are optional, whereas a "+" is a dictation of one or more words, and the asterisk denotes an unconstrained dictation of up to 5 words. In general, we assume that all expansions for *<top_level>* start and end with a silence segment. To prevent spurious action when humans are talking to each other (rather than the robot), we require the presence of an attention word *(<attn>* → "ELI" or "robot") at either the start or end of each such directive.

Fig. 10 shows the result of parsing a simple command to the robot. The boxes indicate the expansion used for each non-terminal category. Notice that there are numerous ways to specify the *<hand_grab>* operation as well as multiple ways to express *<blue>*. After generating a valid parse, the resulting tree of expansions is mined to generate a simple slot-value representation for the utterance (top right). To do this we take each capitalized non-terminal as a slot and assign it the value of whatever first level expansion was used. Thus the whole utterance shown devolves into a request to find blue objects and run the grabbing routine. The majority of the surface words are simply discarded as fillers (e.g. the "*" at the end) or absorbed into irrelevant categories (e.g. *<intro>*). The result of interpreting the sentence "Quickly pick up a dark blue thing, robot" would be exactly the same. This gives some flexibility to the human interface.

**"Eli, please grab the blue bottle now."** → { CMD=hand_grab, COLOR=blue }

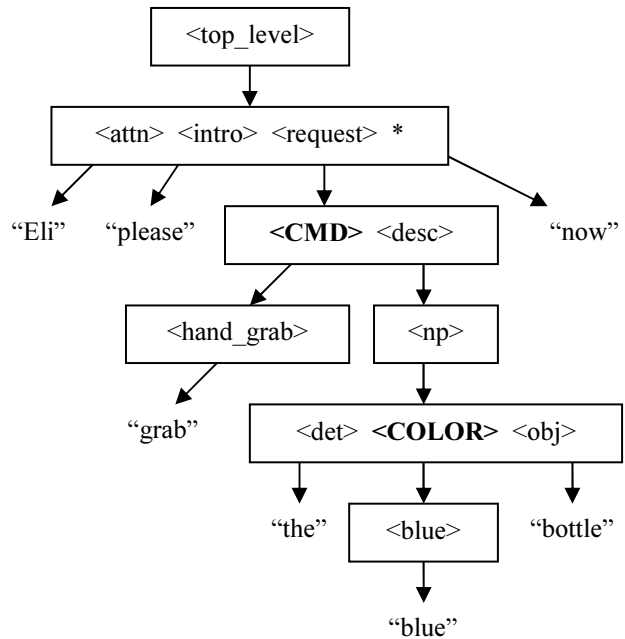

**Fig. 10.** A full utterance is converted to a set of slots and values based on the capitalized categories and their immediate children in the parse tree.

When the robot learns new things it needs to expand its grammar. A simple case is naming of objects, as described in Section 3.2. Here an utterance like "This object is aspirin, Eli" gets parsed under the *<FACT>* branch of *<top_level>* (see Fig. 9) to yield the slot-value representation: {POINT=object, FACT, DICT=aspirin}. To properly process this new object in the future, we add the declared name to the *<NAMES>* category in the grammar. This is kept distinct from generic nouns like "object" because items in the *<NAMES>* class usually have one or more visual models associated with them, and hence invoke a special visual finding routine.

The robot can also learn new actions through verbal interaction, as described in Section 3.4. Fig. 11 shows part of the grammar used to perform this (largely an expansion of <learn> from Fig. 9). Procedure learning is initiated through a set phrase like "Eli, let me show you how to do something" which resolves into {NEW-ACT}. After this, verbal commands are accumulated until a set termination phrase like: "That is how you nudge something, Eli". This resolves into {FINISH, ACT-1=nudge} which cause the robot to stop recording steps and to bind the learned routine to the transitive verb "nudge" (requires 1 argument).

| | | | |
|---|---|---|---|
| =[learn] | =[teach] | =[FINISH] | =[ACT-0] |
| <NEW-ACT> do something | I'm going to | that's how you | wave |
| <NEW-ACT> <ACT-0> | I am going to | that is how you | |
| <NEW-ACT> <ACT-1> <arg> | let me | | =[ACT-1] |
| <FINISH> do it | | | poke |
| <FINISH> <ACT-0> | =[demo] | =[vp] | nudge |
| <FINISH> <ACT-1> <arg> | show | do it | |
| | tell | | |
| =[NEW-ACT] | teach | =[arg] | |
| <teach> <demo> you how to | | something | |
| | | an object | |
| | | <desc> | |

**Fig. 11.** This is a fragment of the grammar the robot uses to learn new verbs. Learning is initiated and terminated by special phrases.

## 2.5 Manipulation Routines

To actually grab an object, the 2D image coordinates must be turned into a 3D position for the arm. To do this we compute a homography based on 4 calibrated points to map 2D image locations to 2D locations on the table surface. This is shown in Fig. 12 where a sheet of printer paper (of known dimensions) has been placed on the table in a known position relative to the arm's coordinate system. The mapping from image coordinates to arm coordinates (at the table's surface only) is indicated by the superimposed purple grid.

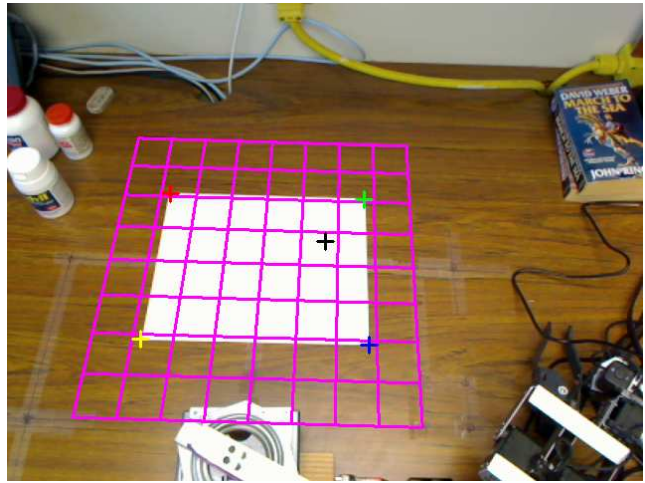

**Fig. 12.** A reference object with known coordinates is used to establish a hand-eye coordination homography.

To find a 3D grasp position for an object, we select its base reference point (as described in Fig. 5) and transform this in order to find the arm coordinates x and y. A fixed z position of 1.5 inches above the table is then specified to complete the grasp point. We also define a "via" point in front of the object to guarantee that the gripper approaches from a reasonable direction. This point is found by converting the object's axis direction into a table surface angle, then picking a spot about 3.5 inches in front of the object's base along this surface angle. To grasp the object, we fully open the gripper and point it along the surface direction. Then we use the inverse kinematics of the arm to plot a linear trajectory from the current position to the computed "via" point, and then another short straight segment to the reach final 3D grasp point.

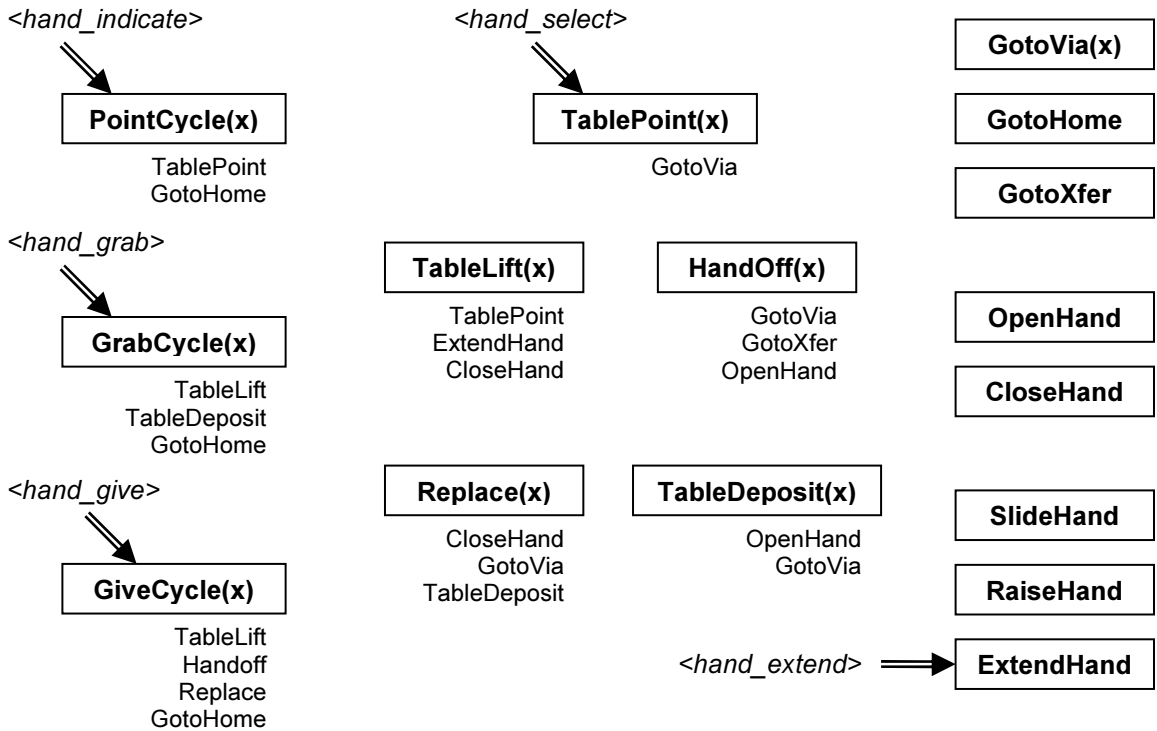

**Fig. 13.** Manipulation is handled by primitive kernel routines (far right) that are incorporated as parts of larger routines. Various grammar categories are mapped directly to invocation of these routines.

The operation and sequencing of these trajectories is controlled by a series of finite state machines (FSMs). There is collection of basic routines, as shown on the far right of Fig. 13, which represent primitive arm moves of various types. More complicated routines are then built up from these as shown to the middle and left, where each routine lists the lower level routines it leverages. While these fancier routines could be learned from interaction (as with "poke", described in Section 3.4) we have hard-coded these particular examples since they are so common. Generally we try to pre-program as much as possible, saving learning for where it is really required.

Being finite state machines, these controllers set some control parameters, then wait for certain sensory conditions to be achieved before moving onto some other set of control parameters. A simple example is the ExtendHand routine. This finds the current position and pointing direction of the hand, plots a linear trajectory to a point about 3 inches further along the current direction, starts movement of the arm, and terminates when the hand gets reasonably close to the specified destination. A more complicated example is CloseHand. Here the gripper is slowly driven toward zero width until a sufficient force is detected. After that, it is widened slightly until the force drops slightly. Other FSMs take one or more arguments, such as GrabCycle(x). This is the basic pickup routine described previously which needs the object position (for TableLift and TableDeposit) and to determine a via point (GrabCycle → TableLift → TablePoint → GotoVia).

Generally, top level verb categories (in angle brackets) are directly linked to one of these FSMs. The simplest example is that if the parsing indicates a command of the form *<hand_extend>* then the ExtendHand FSM is invoked. For routines that take an object as an argument, there must be one unique object, selected either gesturally or by a verbal description. If there is more than one possible binding for this argument (or none) a dialog action is undertaken to try and resolve the situation.

## 3. Demonstration of Abilities

We have described how ELI, our speech-based robot manipulator, selects and moves objects around on a table. We explained how the language parsing works, how objects are found, and how human gestures are detected. The robot is able to answer questions about the scene in front of it and resolve ambiguities in any commands it receives. In addition it can be taught the names of objects and can use these labels to access information in remote databases. Finally, it is also possible to "program" the robot by teaching it new named action sequences. The operation of the system and these components will be illustrated via transcripts from a series of video experiments (Connell 2012) with the actual robot. It is highly recommended to watch this video before reading the rest of the chapter.

## 3.1 Scene Understanding

Using the visual object detection and characterization methods previously described, along with a more complex semantic grammar, the robot can grab objects specified by color, size, position, or gesture. It can also answer questions about objects that have been selected in this way. Fig. 14 provides a transcript of an experiment testing the robot's proficiency.

| | |
|---|---|
| **"Grab it."** (1 object) | |
|     <grabs object> | ← *no confusion since only 1 choice for "it"* |
| **"Grab it."** (4 objects) | |
| "I'm confused. Which of the 4 things do you mean?" | ← *knows a unique target is required* |
| **"What color is the object on the left?"** (4 objects) | |
| "It's blue." | ← *understand positions & colors* |
| **"Grab it"** (4 objects) | |
|     <grabs blue object> | ← *uses "it" from previous interaction* |
| **"Grab that object"** (human points) | |
|     <grabs object> | ← *understands human gesture* |
| **"Grab the white thing."** (2 white objects) | |
| "Do you mean this one?" <robot points> | ← *uses gesture to suggest a choice* |
| **"No, the other one."** | |
|     <grabs other object> | ← *uses "other" from previous interaction* |
| **"Grab the green thing."** (head of lettuce) | |
| "Sorry, that's too big for me." | ← *sensitive to physical constraints* |

**Fig. 14.** As this transcript of one of the video demos (Connell 2012) shows, the robot can resolve pronouns based on context, understand gestures, and request clarification when needed.

One interesting aspect of this conversation is how the robot resolves pronouns through non-linguistic means. If there is only one object present, the binding for "it" is obvious. However if there are several objects, the robot will execute a dialog move to seek clarification. By contrast, if some particular object had recently been mentioned, the robot assumes that this is the proper referent for the pronoun instead. Eli is also capable of executing a mixed mode dialog response, as when it suggests which of the two white objects the user might have wanted by pointing. Finally, the robot also knows the limits of its own abilities in terms of reach and grasping size. That is why, when directed to grasp the green object (the head of lettuce shown in the right panel of Fig. 15), it demurs.

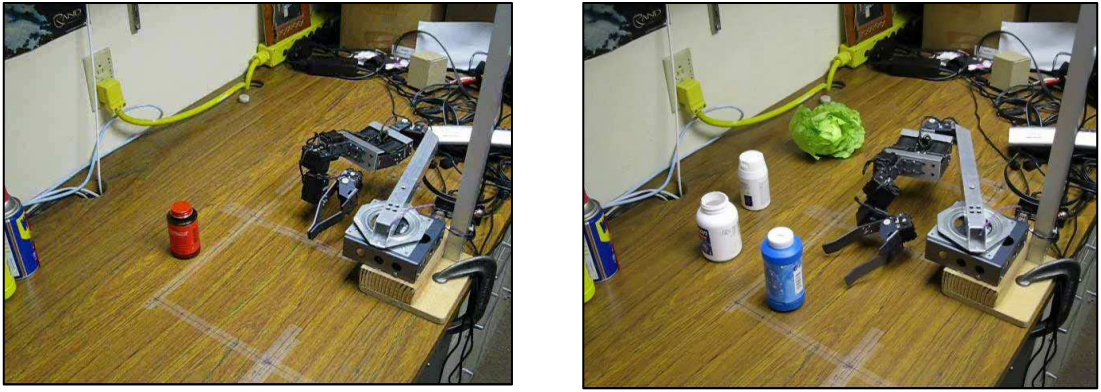

**Fig. 15.** Pronoun resolution depends on shared perception of the scene, not simply previous utterances.

## 3.2 Object Naming

While colors, sizes, positions, and pointing can be used to draw attention to specific objects, in some cases it is more convenient to give objects names. One can then simply say "Give me the WD-40" and have the robot figure out which object this is. Of course to do this, the robot must know that "WD-40" is a valid object. It must also know what the object looks like in order to find it. To teach the robot new nouns like this, we use a simple speech pattern: "NP is called X". Here the NP is any valid noun phrase in the grammar, such as "The big bottle" or "That thing" (with pointing). The X is then either drawn from a list of likely (but unknown) object words, or is an unconstrained dictation item. The actual grammar operations are discussed in Section 2.4.

An interesting problem we have run into is that the dictation results are not always reliable. For instance, when the user says "aspirin" the system sometimes hears "offering". For a speech-only system this is fine since a name is just a random acoustic label. If the robot hears "Pick up the offering" it will perform the correct action. In fact, humans managed to exist for thousands of years with just such cues, having no written language or fixed orthography. However when trying to look up properties of an object elsewhere (as in Section 3.3), the correct term "aspirin" yields much more relevant information.

When the user names an object, the other thing that happens is that a visual model of the object is built. This consists of a coarse size and shape description, plus a histogram of semantic color features (e.g. 50% blue, 30% yellow, 20% red, as described in Fig. 6). For our small universe of objects on a table, this is sufficient to find similar objects. If the same name is taught multiple times, the system will learn multiple models for the object provided that the new view is far enough away from the old representation. These examples are used with a nearest-neighbor classifier to add robustness, since the appearance of objects often varies from side to side, or from different vantage points.

To determine a name for an unknown object, for instance to answer a question like "Eli, what is this object?", we consult the database of visual models. That is, we extract a visual description for all the indicated object, then find the closest stored representation (within a maximum difference limit). The name associated with the best match representation is then taken to be the name of the visible object. The same procedure is used to select an object of a some category as part of a manipulation action, except that here we typically compare all visible objects against a particular given model, like "Tylenol". Note, that although an object can be described verbally with enough specificity to select it from among other items, when the robot actually experiences an object it can build a much richer and more selective model.

---

**"Eli, what is the object on the left?"**
  "I don't know."                                            ← *no existing visual model matches object*
**"Eli, that is aspirin."**
    <new word added to grammar>                    ← *word acquired via dictation*
    <new visual model for object>
"Okay. This is aspirin." <points>
**"Eli, this object is Advil."** (human points)
    <additional visual model for object>          ← *word already known*
  "Okay. That is Advil."
**"Eli, how many Advil do you see?"**
  "I see two."                                                 ← *uses existing visual models to find item(s)*
**"Eli, give me the Tylenol."**
    <gets bottle>                                            ← *uses existing visual model to find item(s)*
  "Here you go"
    <waits for user hand motion> <releases>
    <waits for user hand motion> <regrabs bottle>
"Thanks."
    <replaces bottle>
**"Eli, where is the aspirin?"**
  "Here." <points>                                          ← *uses existing visual model to find item(s)*

---

**Fig. 16.** As this transcript of one of the video demos (Connell 2012) shows, the robot can be taught new nouns by simply showing it objects. The new visual model can then be used in various ways.

Fig. 16 gives the transcript from an experiment in which the robot's learning of new nouns was tested. As can be seen, objects can be indicated either verbally or by pointing. The robot can then use its learned models to find things, count them, and name them when requested.

## 3.3 Semantic Web Access

Many useful functions can be performed in the eldercare domain by robot with just the perceptual and manipulation capabilities already described. However, we can also provide smarter guidance about proposed actions using external data. The IBM Tokyo Research Lab has built a remote consultation agent called Brainy Robot And Intelligence Networked System (BRAINS) that has access to richer semantic information, largely based on the names (types) of objects. Every time the robot interprets a local utterance, we form a potential action plan and transmit this (via TCP/IP socket) to this supervisory program for vetting. A sample of the communication is shown in Fig. 17. As can be seen, the robot generates semantic network triples describing the proposed action. Then the supervisor can either accept or veto the action, or counter-propose some other alternative action.

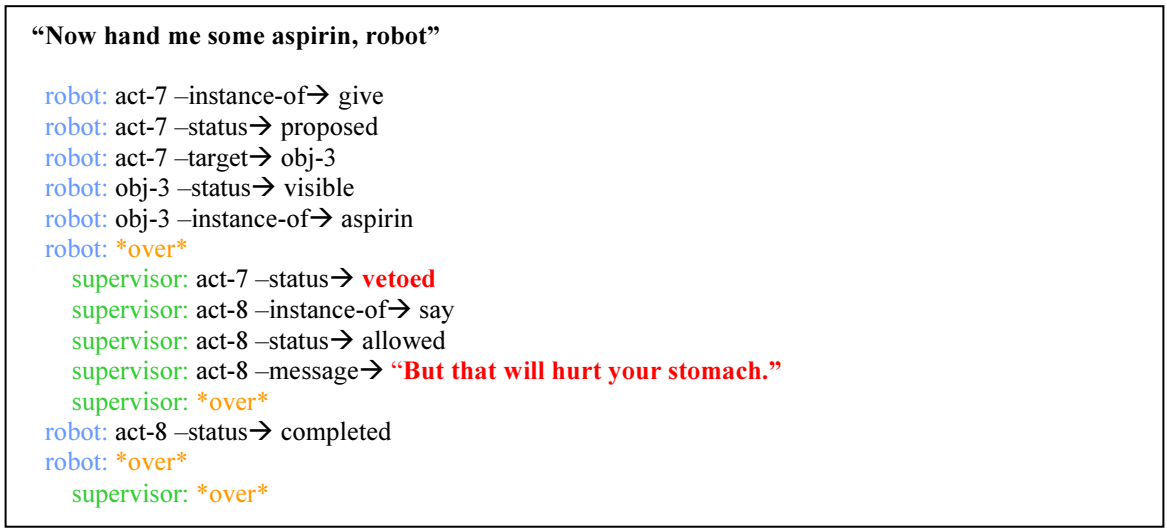

**Fig. 17.** The robot can communicate with a supervisory system using semantic network triples.

The BRAINS supervisor has access to a variety of specialized knowledge as shown in Fig. 18. This includes a database detailing any drug restrictions or interactions for the patient. It also includes a timeline of when certain actions were taken, as well as general background knowledge such as drug taxonomies. Potentially much richer knowledge sources, like RoboEarth (Weibel et al. 2011), could also be employed by the supervisor. This would give the robots a "collective consciousness" where anything one robot learned could potentially be used by all others.

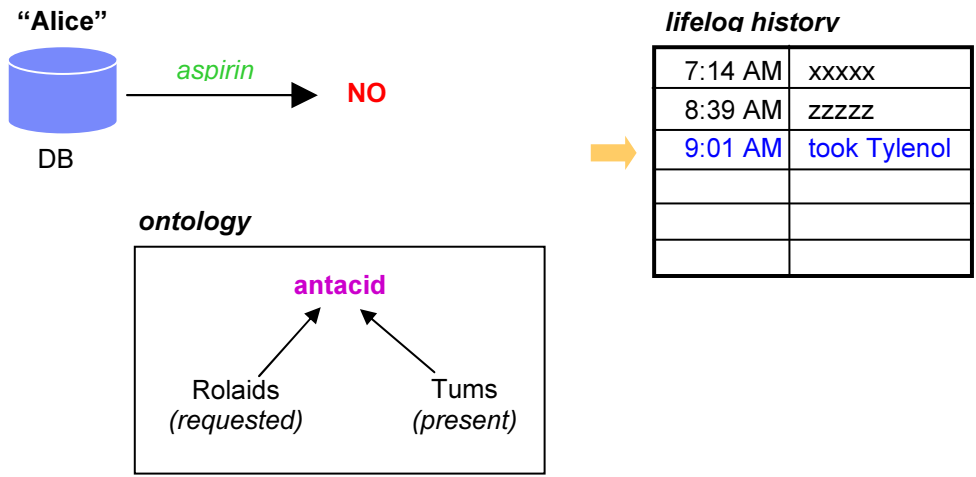

**Fig. 18.** The supervisor has several sources of additional knowledge linked to particular object types.

Fig. 19 gives the transcript of an experiment with the supervisor in the loop. In one case, it consults a database for the user and discovers an aspirin intolerance and thus vetoes dispensing it. Tylenol (paracetamol) does not raise such concerns, hence the supervisor allows this action to be performed. However it records when the Tylenol was given in the personal history (LifeLog) for the user. Thus, when in the last interaction the user again requests Tylenol (perhaps because of memory loss or simply impatience) the supervisor vetoes the action because sufficient time has not elapsed between doses. The other interaction demonstrated here makes use of a taxonomy built for IBM's Jeopardy! project (Murdock et al. 2012) by harvesting WikiPedia entries. The user requests a medication (Rolaids) which is not only unknown, but not present on the table. Yet by using the taxonomy and information about the scene, the supervisor can suggest a similar item that is present.

---

**"Eli, this object is aspirin."** (human points)
    <new word added to grammar>       ← *word acquired via dictation*
    <new visual model for object>
"Okay. That is aspirin"
**"Eli, the object on the right is called Tums."**
    <new visual model for object>     ← *word already known*
"Okay. This <points> is Tums."
**"Eli, give me some aspirin."**
    <check against personal database>     ← *uses existing visual model to find item(s)*
"But that will hurt your stomach."
**"Eli, give me some Tylenol instead."**
    <gets bottle>     ← *uses existing visual model to find item(s)*
"Here you go"
    <waits for user hand motion> <releases>
    <waits for user hand motion> <regrabs bottle>
"Thanks."
    <replaces bottle> <records dose in lifelog>
**"Eli, give me some Rolaids."**
 "I don't know what Rolaids looks like."     ← *no visual model for item*
    <ontology used to find available alternative(s)>
"Do you want another antacid, Tums?"
**"Eli, just give me some Tylenol."**
    <lifelog consulted for last dose>     ← *uses existing visual model to find item(s)*
"You just had Tylenol."

---

**Fig. 19.** As this transcript of one of the video demos (Connell 2012) shows, the backed system can look up personal information, reason about substitutions, and monitor events over time.

### 3.4 Procedure Learning

Not only can Eli learn new nouns, he can also learn verbs. Fig. 20 shows the transcript from an experimental run where the robot is being taught to poke things. The user teaches the action as a series of steps, like a verbal scripting language, as opposed to imparting some declarative specification of a desired result state. The steps themselves are indexical (as needed), so when they are composed, the whole sequence is also indexical. In other words, since the component "point" action requires a focus object, the resulting "poke" action does also. As the later part of the transcript indicates, once an action has been learned it can be directly applied to other objects in the scene using the base grammar.

**"Eli, poke the thing in the middle."**
    <new action sequence opened for input>        ← *no existing action sequence to link*
"I don't know how to poke something."
**"Eli, point at it."**
    <points>        ← *resolves pronoun from previous selection*
**"Eli, extend your hand."**
    <advances>        ← *low level incremental move*
**"Eli, retract your hand."**
    <retreats>        ← *low level incremental move*
**"Eli, that is how you poke something."**
    <links action sequence to word>        ← *recognizes closing of action block*
"Okay. Now I know how to now poke something."
**"Eli, poke the red object."**
    <pokes>        ← *retrieves action sequence for verb and executes*
**"Eli, poke the object on the left."**
    <pokes>        ← *retrieves action sequence for verb and executes*
**"Eli, poke the Tylenol."**
    <pokes>        ← *retrieves action sequence for verb and executes*

**Fig. 20.** As this transcript of one of the video demos (Connell 2012) shows, the robot can be taught a new verb by simply walking it through the appropriate procedural steps.

Learning a new verb requires both adding the trigger word to the grammar and binding it to a new action routine. Much of this has already been discussed in relation to Figs. 11 and 13. The resulting new FSM for "poke" is shown on the right of Fig. 21. This "macro" sequence is now invoked when the word "poke" is used as a verb. And, since the user can call for it directly, this routine can also be included as a step in some other more complicated learned procedure.

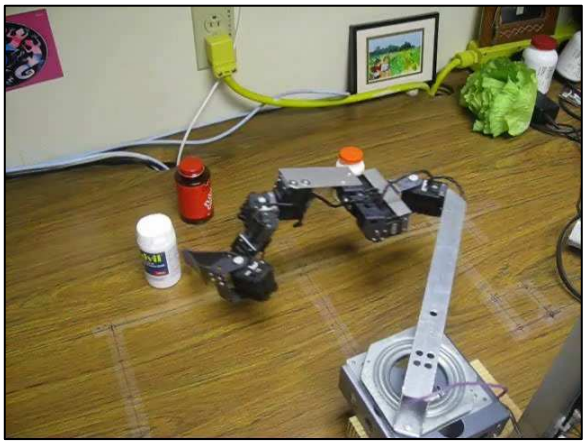

**"poke"**

| TablePoint | 1.0 |
| ExtendHand | 1.0 |
| ExtendHand | -1.0 |

**Fig. 21.** The robot learns how to "poke" something as a parameterized sequence of actions.

## 4. Adding motivation

There is a feeling of a somewhat eerie presence when interacting with a system such as Eli. It responds to your commands to perform real-world actions, answers questions, and discusses things with you when unsure. Yet this system is still a "slave"; what it lacks to make it feel truly sentient is self-motivation. This is not to say that AI's should run off and do their own thing in a corner somewhere (or worse, try to take over the world). Instead, anyone who has ever had an assistant knows that the most valuable ones anticipate needs and questions, and act of their own volition to find answers to sub-problems that may arise. The question is how to imbue computers with a similar facility, not necessarily to make them autonomous, but to make them more useful.

In light of this, it is instructive to examine why animals do what they do and when they do it. Importantly, unlike Eli, animals do things spontaneously all the time. Pets do not await a command from their owners, and wild animals obviously have no external agency calling the shots. In this way they are very different from contemporary computer programs which amplify and elaborate on imperatives supplied by their users. Moreover, although goals can be internally generated rather than supplied from the outside, it seems unlikely that animals are always operating in a goal-driven manner, except in the loosest sense. That is, much of what animals seem to do is routine or reflexive – there is no specific articulated goal which they are pursuing. Ongoing activities such as foraging, grooming, and migration can not be traced back through a logical chain of reasoning to a concrete goal proposition.

To mirror this observation, the motivation system proposed here has only a loose coupling between goals and actions. As shown in Fig. 22, the bulk of the activity is controlled reactively using situation-action policies (Brooks 1986, Connell 1990). Some of these policies are active all the time, while others can be switched in or out depending on the situation (Connell & Viola 1990). Much of the switching of polices is governed by a set of persistent directives. In this respect the action component of the architecture is similar to Minsky's K-lines (Minsky 1979).

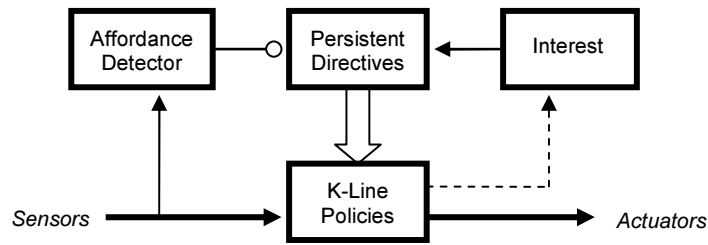

**Fig. 22.** A loosely coupled architecture for self-motivation could greatly enhance the utility of robots.

Yet where do these policies and directives come from? One possible answer is that we can exploit <S, E, A> triples. Here E is an "exciting" event, S represents the observed situational context, and A captures the current actuator settings. The rules governing their formation and use are summarized below.

$$S \,\&\, E \,\&\, A \,\&\, I(E) \quad \rightarrow <S, E, A>$$

$$<S, E, A> \,\&\, S \,\&\, I(E) \quad \rightarrow D(E)$$

$$D(E) \,\&\, <S, E, A> \,\&\, S \quad \rightarrow A$$

$$D(E) \,\&\, <S, E, A> \quad \rightarrow I(S)$$

These <S, E, A> triples are not recorded all the time, but only when something "interesting" happens (the operator I(E) in the first line). Once recorded, there are three uses for <S, E, A> triples. First they can function as affordance detectors – determining when the environment may be offering a useful opportunity to the agent. In this case the recorded S is used to predict the interesting additional observation E that might already hold, or may be forced to become present through some action (second line). Although E was originally "interesting" when it was recorded, the organism's desires and needs routinely change over time. Thus each such proposed affordance is evaluated against the current "interest" metric (I). Then, if sufficiently stimulating, a desire (D) for this event is latched in as one of the persistent directives that control the behavioral policies.

The other major use for <S, E, A> triples is as part of a policy. If the E part matches one of the current directives then when situation S occurs the agent will be prompted to try action A (third line). Triples can also be used for loose backwards chaining. If E is one of the active directives then the agent should become interested in situation S (fourth line). In this way it can serendipitously learn how to accomplish subgoals in the course of an activity (whether or not the associated action is taken).

As an example, suppose you are walking along the shore of a pond and there is a sudden splash as a frog jumps into the water. Since sudden noises are intrinsically interesting this would prompt the formation of a triple like <pond, splash, walk-along>. Now every time you see a pond the splash will be brought to mind. If this is still an interesting occurrence you will latch it in as something you desire to happen. This in turn activates one or more policies that bias you into doing things that might lead to a splash, such as if you see a pond you should walk along its shore. The desire for splashing may also activate other triples such as <pond & rock, splash, drop-rock>. Since the situation part of this is not yet fulfilled its action is not performed. Yet the set of interesting things will be expanded temporarily to include rocks, prompting background learning about where to find rocks (for this or any other purpose).

In some sense the formula for AI may well be 50% robotics, 40% language, and 10% motivation. With ELI we have the first two bases covered; it is only some form of this final motivation component that is needed to complete the picture.

**References**

(Anderson 1983) J. R. Anderson, <u>The Architecture of Cognition</u>, Lawrence Erlbaum, 1983.

(Breazeal et al. 2006) C. Breazeal et al., "Using perspective taking to learn from ambiguous demonstrations", Robotics and Autonomous Systems, 54, pp. 385-393, 2006.

(Brooks 1986) R. Brooks, "A Robust Layered Control System for a Mobile Robot", *Journal of Robotics and Automation*, RA-2, pp. 14-23, 1986.

(Chitta et al. 2012) S. Chitta, E. Jones, M. Ciocarlie, and K. Hsiao, "Perception, Planning, and Execution for Mobile Manipulation in Unstructured Environments", *IEEE Robotics and Automation Magazine*, 19(2), pp. 58-71, June 2012.

(Choi et al. 2009) Y. Choi et al., "Hand It Over or Set It Down: A User Study of Object Delivery with an Assistive Mobile Manipulator", *IEEE Int. Symp. on Robot and Human Interactive Communication (RO-MAN)*, pp. 736-743, 2009.

(Connell 1990) J. Connell, <u>Minimalist Mobile Robotics</u>, Academic Press, 1990.

(Connell 2012) J. Connell, "ELI Robot Arm Demos", online video, <u>http://www.youtube.com/watch?v=M2RXDI3QYNU</u> or <u>http://www.johuco.com/eli_arm_demos.wmv</u>, 2012.

(Connell & Viola 1990) J. Connell and P. Viola, "Cooperative Control of a Semi-Autonomous Mobile Robot", *Proc. of the IEEE Conf. on Robotics and Automation (ICRA-90)*, pp. 1118-1121, 1990.

(Lopes & Teixeira 2000) L. Seabra Lopes and A. Teixeira, "Human-Robot Interaction through Spoken Language Dialogue", *IEEE Int. Conf. on Intelligent Robots and Systems (IROS)*, pp. 528-534, 2000.

(Minksy 1979) M. Minksy, "K-Lines: A Theory of Memory", MIT AI Memo 516, 1979.

(Murdock et al.) J. Murdock et al., "Typing candidate answers using type coercion", *IBM J. of Res. and Dev.*, 56(3/4), pp. 7:1-13, 2012.

(Peltason et al. 2009) J. Peltason et al., "Mixed-Initiative in Human Augmented Mapping", *IEEE Int. Conf. on Robotics and Automation (ICRA)*, pp. 2146-2153, 2009.

(Pryor 1984) K. Pryor, <u>Don't Shoot the Dog</u>, Simon & Schuster, 1984.

(Roy 2003) D. Roy, "Grounded Spoken Language Acquisition: Experiments in Word Learning", *IEEE Trans. on Multimedia*, 5(2), pp. 197-209, 2003.

(Siskind 2001) J. Siskind, "Grounding the Lexical Semantics of Verbs in Visual Perception using Force Dynamics and Event Logic", *J. of Artificial Intelligence Research*, 15, pp. 31-90, 2001.

(Skinner 1957) B. F. Skinner, <u>Verbal Behavior</u>, Appleton-Century Crofts, 1957.

(Soltau et al. 2010) H. Soltau, G. Saon, and B. Kingsbury, "The IBM Attila Speech Recognition Toolkit", *IEEE Spoken Language Technology Workshop (SLT)*, pp. 97-102, 2010.

(Srinivasa et al. 2010) S. Srinivasa et al., "HERB: A Home Exploring Robotic Butler", *Autonomous Robots*, 28(1), pp. 5-20, 2010.

(Steels 1998) L. Steels, "The origins of syntax in visually grounded robotic agents", *Artificial Intelligence*, 103, pp. 133-156, 1998.

(Stuckler et al. 2012) J. Stuckler, D. Holz, and S. Behnke, "Demonstrating Everyday Manipulation Skills in RoboCup@Home", *IEEE Robotics and Automation Magazine*, pp. 34-42, June 2012.

(Vygotsky 1962) L. Vygotsky, <u>Thought and Language</u>, MIT Press, 1962.

(Weibel et al. 2011) M. Weibel et al., "RoboEarth – A World Wide Web for Robots", *IEEE Robotics and Automation Magazine*, 18(2), pp. 69-82, June 2011.